\documentclass[twoside]{article}
\usepackage{amsfonts,amssymb,amsbsy,textcomp,marvosym,amsmath,caption,threeparttable,amsthm}
\usepackage{eurosym,mathrsfs,fancyhdr,CJK,multicol,graphics,indentfirst,color,bm,upgreek,booktabs,graphicx}
\usepackage{gensymb}
\usepackage{float}
\usepackage{graphicx}
\usepackage[caption=false]{subfig}
\usepackage[titletoc]{appendix}

\newcommand{\N}{\mathbb{N}}
\newcommand{\Z}{\mathbb{Z}}
\newcommand{\mini}[1]{\begin{minipage}[c]{0.14\linewidth}{
	\includegraphics[width=\textwidth,height=\textwidth,keepaspectratio]{#1}
}\end{minipage}}
\newcommand{\huo}[1]{{\textcolor[rgb]{0.0,0.0,0.5}{\emph{[Huo: #1]}}}}
\newcommand{\Skip}[1]{}

\newcommand{\sfsnet}[1]{
	\mini{#1input_face.png}
	\mini{#1cropped_sfsnet_reconstruction.png}
	\mini{#1cropped_sfsnet_albedo.png}
	\mini{#1cropped_sfsnet_shading.png}
	\mini{#1cropped_sfsnet_normal.png}
	\mini{white.png}	
	\rotatebox[origin=c]{270}{\footnotesize SfSNet}
}

\newcommand{\our}[1]{
	\mini{white.png}
	\mini{#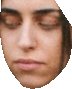}
	\mini{#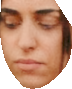}
	\mini{#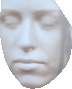}
	\mini{#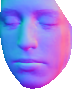}
	\mini{#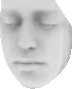}
	\rotatebox[origin=c]{270}{\footnotesize Our}
}

\newcommand{\relightinghumans}[1]{
	\mini{#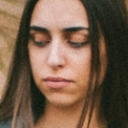}
	\mini{#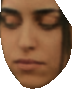}
	\mini{#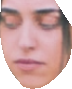}
	\mini{#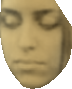}
	\mini{white.png}
	\mini{white.png}
	\rotatebox[origin=c]{270}{\footnotesize ReH}
}

\newcommand{\beard}[1]{
	\mini{#1input_face.png}
	\mini{#1cropped_inf_albedo.png}
	\mini{#1cropped_inf_visibility.png}
}

\newcommand{\sfsnettransfer}[1]{
	\mini{#1source.png}
	\mini{#1source-sfsnet-cropped-shading.png}
	\mini{#1transfer-sfsnet-cropped-shading.png}
	\mini{#1transfer-sfsnet-cropped-rendering.png}
	\mini{#1target-sfsnet-cropped-shading.png}
	\mini{#1target.png}
	\rotatebox[origin=c]{270}{\footnotesize SfSNet}
}

\newcommand{\ourtransfer}[1]{
	\mini{white.png}
	\mini{#1source-cropped-shading.png}
	\mini{#1transfer-cropped-shading.png}
	\mini{#1transfer-cropped-rendering.png}
	\mini{#1target-cropped-shading.png}
	\mini{white.png}
	\rotatebox[origin=c]{270}{\footnotesize Our}
}

\newcommand{\componentcaption}{
	\vspace{1mm}
	\begin{minipage}{\textwidth}
		\footnotesize{\quad\, Input \;\;\; Recon. \; Albedo \; Shading \; Normal \; Visibility}
	\end{minipage}
}

\def\footnoterule{\kern 1mm \hrule width 10cm \kern 2mm}

\allowdisplaybreaks
\catcode`@=11
\def\title#1{\vspace{3mm}\begin{flushleft}\vglue-.1cm\Large\bf\boldmath\protect\baselineskip=18pt plus.2pt minus.1pt #1
\end{flushleft}\vspace{1mm} }
\def\author#1{\begin{flushleft}\normalsize #1\end{flushleft}\vspace*{-4pt} \vspace{3mm}}

\catcode`@=11
\def\section{\@startsection{section}{1}{\z@}%
 {-3ex \@plus -.3ex \@minus -.2ex}%
 {2.2ex \@plus.2ex}%
{\normalfont\normalsize\protect\baselineskip=14.5pt plus.2pt minus.2pt\bfseries}}
\def\subsection{\@startsection{subsection}{2}{\z@}%
 {-3ex\@plus -.2ex \@minus -.2ex}%
 {2ex \@plus.2ex}%
{\normalfont\normalsize\protect\baselineskip=12.5pt plus.2pt minus.2pt\bfseries}}
\def\subsubsection{\@startsection{subsubsection}{3}{\z@}%
 {-2.2ex\@plus -.21ex \@minus -.2ex}%
 {1.4ex \@plus.2ex}
{\normalfont\normalsize\protect\baselineskip=12pt plus.2pt minus.2pt\sl}}

\begin{document}
\begin{CJK*}{GBK}{song}
\thispagestyle{empty}
\vspace*{-13mm}
\vspace*{2mm}

\title{Normal and Visibility Estimation of Human Face from a Single Image}

\author{Fuzhi Zhong, Rui Wang, Yuchi Huo, Hujun Bao}

\let\thefootnote\relax\footnotetext{{}\\[-4mm]\indent\ Regular Paper}

\noindent {\small\bf Abstract} \quad  {\small Recent work on the intrinsic image of humans starts to consider the visibility of incident illumination and encodes the light transfer function by spherical harmonics. In this paper, we show that such a light transfer function can be further decomposed into visibility and cosine terms related to surface normal. Such decomposition allows us to recover the surface normal in addition to visibility. We propose a deep learning-based approach with a reconstruction loss for training on real-world images. Results show that compared with previous works, the reconstruction of human face from our method better reveals the surface normal and shading details especially around regions where visibility effect is strong.}

\vspace*{3mm}


\vspace*{4mm}

\end{CJK*}
\baselineskip=18pt plus.2pt minus.2pt
\parskip=0pt plus.2pt minus0.2pt
\begin{multicols}{2}

\begin{figure*}[tbp]
	\centering
	\mbox{} \hfill
	\includegraphics[width=\linewidth]{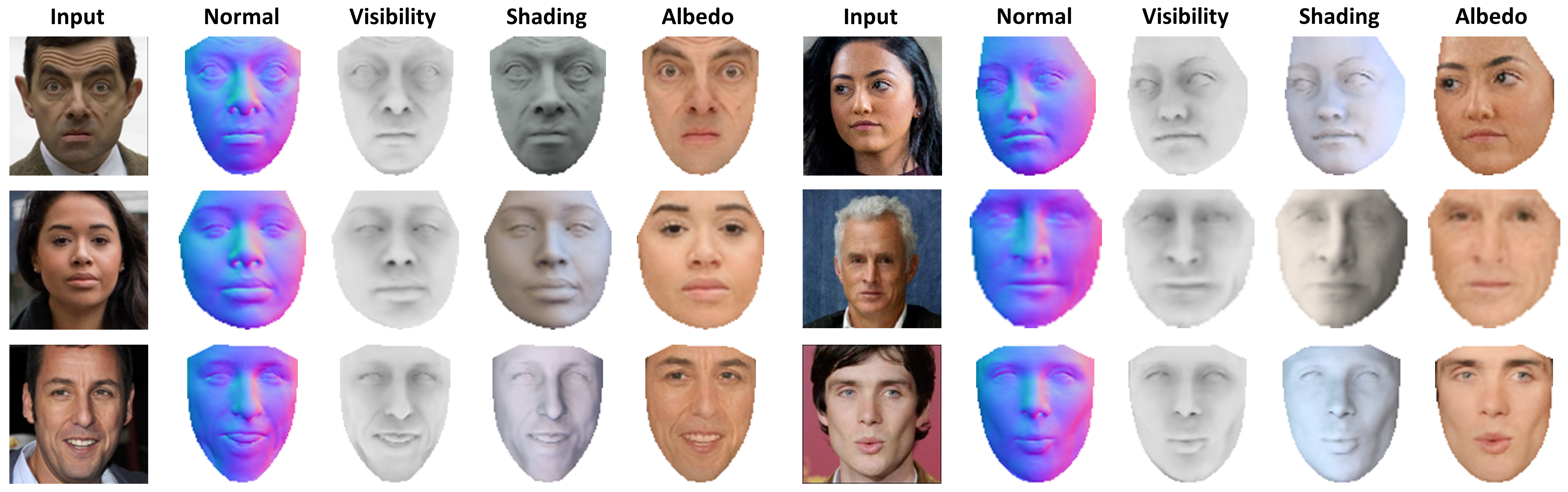}
	\hfill \mbox{}
	\caption{
		Given a single human face image, our method decomposes it into surface normal, visibility, illumination and albedo. The illumination, surface normal and visibility together produce shading. The dark regions in the image are successfully modeled by our visibility term. By separating the light transfer function into surface normal and visibility, we obtain better normal and shading estimation where visibility effect is strong. In our paper, visibility functions are represented by spherical harmonics. We visualize the ambient term to illustrate visibility in this figure.
	}
\end{figure*}

\section{Introduction}

Understanding face images is of great importance to the computer vision and graphics community, and has been extensively studied. Previous works on the intrinsic decomposition of human face \cite{1,2} or the inverse rendering of human face \cite{3,4,5} assume Lambertian reflectance but neglect the visibility effect. As a consequence, the reconstruction results usually have unnatural brightness in concave regions. Recently `Relighting Humans' \cite{6} is the first intrinsic decomposition method to consider the visibility effect that trains on synthetic data to estimate the light transfer function, which is the product of visibility and cosine functions. However, such work only learns the light transfer function as a whole and cannot infer shape attributes such as surface normal. Additionally, it cannot train on real-world data since they found it difficult to fine-tune on real-world data due to the large degrees of freedom (9 dimensions per pixel) of the inferred light transfer function, thus the reconstruction error of their method on real-world images is large. 
\par
In this paper, we propose a novel intrinsic decomposition method that takes a single human face image as the input, and estimates the surface normal, visibility, incident illumination, and albedo. See Fig.1 for an illustration of this decomposition. We show that the light transfer function can be further decomposed into visibility function and cosine function related to surface normal. We design a novel differentiable rendering layer in a neural network that consists of the integration of the product of illumination, cosine, and visibility functions. With this newly designed neural network, we are able to separate the light transfer function into surface normal and visibility. We propose a more stable reconstruction loss for training real-world images, even with large degrees of freedom in the visibility term. We achieve better reconstruction quality than `Relighting Humans' \cite{6} on the real-world test data they provide. We compare with the state-of-the-art method SfSNet \cite{2} quantitatively on the Photoface dataset \cite{7} and qualitatively on the CelebA dataset \cite{8} to show the correctness of our decomposition. Compared to SfSNet, our method better recovers the surface normal details and produces better shading where visibility effect is strong.
\par
In summary, our main technical contributions are: (1) We decompose the light transfer function
into visibility function and cosine function related to surface normal and design a novel differentiable rendering layer in neural networks, which allows us to recover the surface normal in addition to visibility. (2) We propose a more stable reconstruction loss for training on real-world images, even with large degrees of freedom in the visibility term. Thus our method achieves better reconstruction quality on real-world images.

\section{Related Work}
\textbf{Face reconstruction} from a single image is a highly challenging and ill-posed inverse problem. To deal with this ill-posedness, researchers [9-16] have made additional prior assumptions, such as constraining faces to lie in a low-dimensional subspace, e.g., 3D Morphable Model (3DMM) \cite{17} learned from face scanning. Most of the previous works assume local Lambertian reflectance but neglect global illumination effects. Recently the work~\cite{14} determines the visibility function in each vertex of 3DMM to model visibility effect. But it is limited in their low-dimensional 3DMM subspace, which degrades performance in real-world settings. Our work is inspired by such a work~\cite{14} considering visibility effect but is further formulated as an intrinsic decomposition for real-world human faces by training on mixtures of real-world and synthetic images.
\par
\noindent \textbf{Intrinsic decomposition} \cite{18} proposes to decompose an image into per-pixel physical intrinsic components such as shape and shading. The shape is usually represented by per-pixel surface normal instead of a 3D mesh. Recent work of SIRFS \cite{19} extended this decomposition with low-frequency illumination and extensive priors on shape, albedo and illumination. This formulation has also been used in several works~\cite{1,2,20} on human faces. Recently, Kanamori and Endo~\cite{6} considered the visibility effect and learns the light transfer function. Since they directly estimate the light transfer function as a whole, their method doesn't produce surface normal. While we show that such a light transfer function can be further decomposed into visibility function and cosine function related to surface normal, thus we can recover the surface normal. We employ a `pseudo supervision' training strategy similar to work~\cite{1,2} to achieve better reconstruction quality on real-world test data.
\par
\noindent \textbf{Deep learning for intrinsic decomposition} tries to tackle the intrinsic decomposition, a highly ill-posed problem, by learning from large amounts of data. But the challenge is at collecting ground truth labels for real-world images. Recent approaches for intrinsic decomposition rely on synthetic data generated by physically-based realistic rendering, and utilize in-network differentiable rendering to reconstruct the original image. Deschaintre et al.~\cite{21} used an auto-encoder network for the intrinsic decomposition of a planar surface with spatially-varying reflectance. Li et al.~\cite{22} worked on general objects with spatially varying reflectance and approximated global illumination effect by training a 2D convolutional network as the rendering layer. Sun et al.~\cite{23} learned the relighting function by training a fully convolutional network on real facial data. We follow this trend to learn the intrinsic decomposition and design a novel differentiable rendering layer in neural networks, which explicitly decomposes the light transfer function into visibility and cosine term related to surface normal.

\begin{figure*}[tbp]
	\centering
	\mbox{} \hfill
	\includegraphics[width=\linewidth]{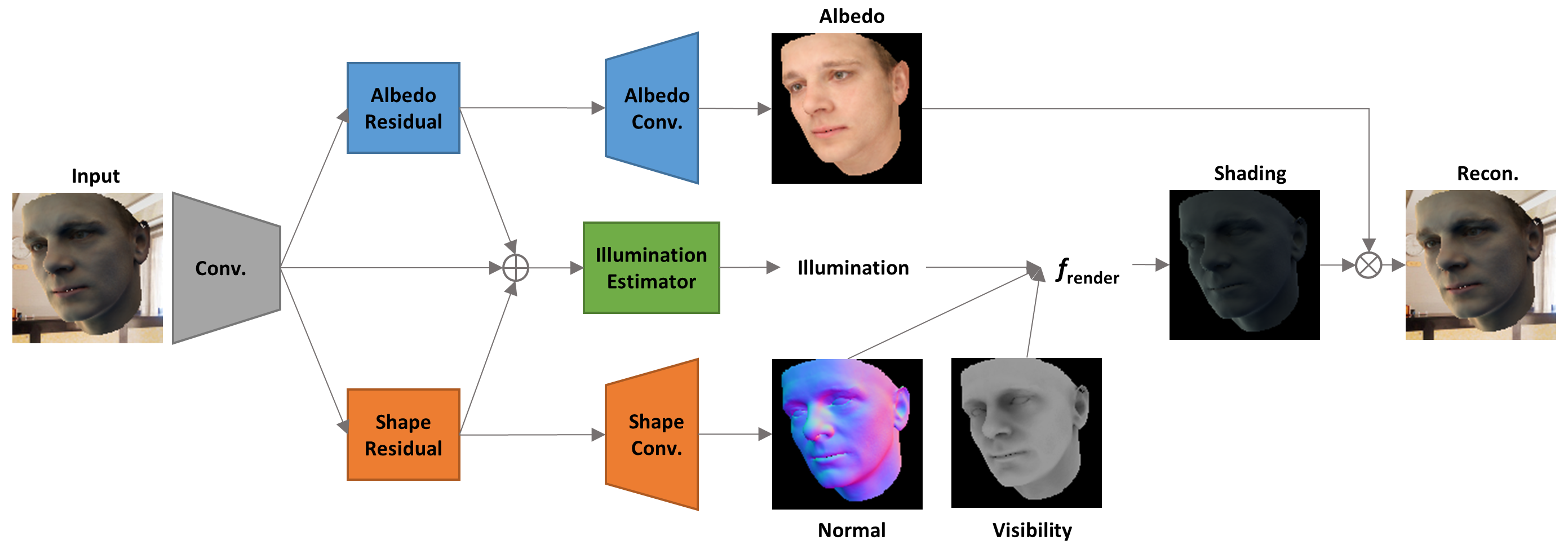}
	\hfill \mbox{}
	\caption{
		Our network consists of residual blocks to extract albedo and shape features. These features are further concatenated with image features to estimate illumination. The illumination, surface normal, and visibility are passed through a differentiable rendering layer to produce shading. The shading and albedo are multiplied together to reconstruct the input image.}
\end{figure*}

\section{Approach}
\subsection{Differentiable Rendering}
We assume the real-world human face image $\hat{I}$ as a rendered result of Lambertian reflectance. Our goal is to recover incident illumination, surface normal, albedo, and visibility of each pixel in $\hat{I}$. Utilizing recovered factors, the original image $\hat{I}$ can be accurately rendered by a forward rendering process. The rendering equation~\cite{24} for each pixel under Lambertian reflectance is defined as:
\begin{equation}\label{eq:f_render}
\begin{aligned}
I &= f_{render}(\rho, n, v, L)\\
&= \frac{\rho}{\pi}\int_{\Omega^+}{L(\omega)T(\omega, n)d\omega}\\
&= \frac{\rho}{\pi}\int_{\Omega^+}{L(\omega)v(\omega)\cos(\omega, n)d\omega},
\end{aligned}
\end{equation}
where $n$ is the surface normal, $\Omega^+$ is the upper hemisphere centered around the surface normal, $\omega$ is the direction of incident illumination, $0 \leq \rho \leq 1$ is the albedo, $L(\omega)$ is the incident illumination function and $T(\omega, n)$ is the light transfer function. We decompose the light transfer function $T(\omega)$ into visibility function $v(\omega)$ and cosine function $\cos(\omega, n)$ related to surface normal, this allows us to recover surface normal in addition to visibility. The visibility function $v(\omega)$ is defined as:
\begin{equation}\label{eq:visibility_function}
v(\omega)= 
\begin{cases}
0,& \text{if } \omega \text{ intersects any object},\\
1,& \text{otherwise}.
\end{cases}
\end{equation}
We assume distant illumination, and represent $L(\omega)$, $v(\omega)$ and $\cos(\omega, n)$ by 2nd degree spherical harmonic coefficients to enhance the integration speed, where the coefficients are obtained by projecting them onto spherical harmonic basis $\{Y_i(\omega)\}_{i=1}^9$ using equation (\ref{eq:sh_projection}):
\begin{equation}
L(\omega) = \sum_{i=1}^{9} L_i Y_i(\omega),
\end{equation}
\begin{equation}
v(\omega) = \sum_{i=1}^{9} v_i Y_i(\omega),
\end{equation}
\begin{equation}\label{eq:cosine_expansion}
\cos(\omega, n) = \sum_{i=1}^{9} c_i Y_i(\omega).
\end{equation}
Note for $\cos(\omega, n)$, we first analytically derive its formula around positive Z-axis:
\begin{equation}
\begin{aligned}
&\cos(\omega, (0,0,1))\\
=&\frac{\sqrt{\pi}}{2} Y_1(\omega) + \sqrt{\frac{\pi}{3}}Y_3(\omega) + \frac{\sqrt{5\pi}}{8}Y_7(\omega),
\end{aligned}
\end{equation}
then rotate it to surface normal $n$ using equation (\ref{eq:sh_rotation}).
Finally the differentiable rendering result in equation (\ref{eq:f_render}) is evaluated using equation (\ref{eq:triple_product_integration}).

\subsection{Training Overview}\label{sec:trainingoverview}

We rely on supervised training, which is the machine learning task of learning a function that maps an input image $\hat{I}$ to the outputs $\{\rho, n, v, L\}$. Since it is difficult to acquire large amounts of training data from the real world, we use computer graphics to generate synthetic images with ground truth labels. We generate 3D Morphable Model (3DMM) \cite{17} in various shapes, reflectance and render them in various illumination and viewpoints. However, training on synthetic data alone cannot well reconstruct real-world images. We further train on the real-world dataset CelebA \cite{8} using reconstruction losses to improve reconstruction.\par

The architecture of our network is illustrated in Fig.2. The network takes the input image to output several components, which are then passed through the differentiable rendering layer to reconstruct the input image. It has been shown in previous works~\cite{2, 25} that low-frequency variations learned from synthetic data can be used as priors or `pseudo supervision' to prevent trivial solutions when training on real-world data whose ground truth labels are unavailable. Our training strategy falls into this paradigm that consists of 2 stages.
In the first stage, we train our network on synthetic data. Then we apply the trained network on real-world data to obtain the estimates. These estimates will be used as pseudo ground truth labels in the second stage. In the second stage, we train on mixtures of synthetic data with ground truth labels and real-world data with the pseudo ground truth labels. Our network architecture is similar to the network used in work~\cite{2}, but with a different differentiable rendering layer, the network details are provided in Section \ref{section:network_architecture_details}.
Our loss function is defined as:
\begin{equation}\label{eq:loss_function}
E = E_{recon} + E_\rho + \lambda_n E_n + \lambda_v E_v + \lambda_L E_L,
\end{equation}
where $E_\rho$, $E_n$, $E_v$ and $E_{recon}$ denote the losses over all pixels in the face region for albedo, surface normal, visibility and reconstruction respectively. $E_{recon}$ is the error between the input image $\hat{I}$ and the image $I$ rendered by equation (\ref{eq:f_render}). Instead of using the error between spherical harmonic coefficients, $E_L$ is defined as the loss of illumination values of uniformly sampled directions on the sphere. $\lambda_n$, $\lambda_v$, and $\lambda_L$ are the balancing weights of the surface normal, visibility and illumination, respectively. For the face region mask of real-world images, we detect 68 facial landmarks using \cite{26} and create a convex hull based on these landmarks.

\section{Implementation Details}
\subsection{Synthetic Data Generation}
We use 3D Morphable Model (3DMM) \cite{27} as the parametric model and manually remove the oral cavity \huo{remove?} to generate various shapes and reflectance. Suppose the mean face is facing forward positive Z-axis with top points to positive Y-axis. We generate random 3DMM, rotate it around Z-axis (roll), X-axis (pitch) and Y-axis (yaw) with angles uniformly distributed in $[-30\degree, 30\degree]$, $[-20\degree, 20\degree]$ and $[-30\degree, 30\degree]$ respectively, then scale and shift its bounding box to fit in $[-1, 1]$.
\par
We collect 203,992 illumination conditions from 3 different dataset \cite{2, 28, 29} to render synthetic data. We project a panorama image of \cite{28} onto a spherical harmonic basis to obtain spherical harmonic coefficients using equation (\ref{eq:sh_projection}). Note that the spherical harmonic coefficients in \cite{2, 29} are shading, so we establish an over-determined system of linear equations to recover the incident illumination (see Appendix \ref{section:recover_illumination_from_shading}). Some illumination in \cite{2, 29} contains negative value on the sphere. We clamp them to be non-negative. We further do data augmentation by rotating the illumination. The overall intensity of the illumination is scaled to make the largest intensity of all pixels in the face region uniformly distributed in $[0.8, 1]$.
\par
We use OpenGL \cite{30} to render the face region mask, surface position, normal, and albedo images. We use full perspective projection and place the camera at $(0,0,5.8)$ pointing to negative Z-axis with horizontal field of views uniformly distributed in $[20\degree, 25\degree]$. For the visibility function in equation (\ref{eq:visibility_function}), we sample 872 uniformly distributed directions on sphere then project onto spherical harmonic basis using equation (\ref{eq:sh_projection}) to get the coefficients $\{v_i\}_{i=1}^9$. We use Optix \cite{31} to accelerate ray intersection in visibility testing. We erode the face region mask inner by one pixel to ignore the loss in equation (\ref{eq:loss_function}) around the boundary of the face region. We use indoor images without human faces to fill the pixels outside the face region. A total of 200,000 synthetic data are generated in 110 hours.

\subsection{Network Architecture}\label{section:network_architecture_details}
The network architecture is shown in Fig.2. Our input, albedo, surface normal, and visibility images are of spatial resolution $128 \times 128$. Let LReLU denotes Leaky ReLU with 0.2 slope \cite{32}. Below we describe the details of each block in the network.\par
\noindent \textbf{`Conv.'}: C64(k7) - C128(k3) - C*128(k3)\\
`CN(kS)' denotes 2D convolutional layers with N $S \times S$ filters with stride 1, followed by Batch Normalization \cite{33} and LReLU. `C*N(kS)' denotes `CN(kS)' with stride 2 without Batch Normalization. The output of `Conv.' produces a tensor of spatial resolution $64 \times 64$.\par
\noindent \textbf{`Albedo Residual'}: 5 ResBLK - BN - LReLU\\
This block consists of 5 residual blocks (ResBLK) \cite{34} followed by Batch Normalization(BN) and LReLU. Each `ResBLK' consists of BN - LReLU - C128 - BN - LReLU - C128. The output of `Albedo Residual' produces a tensor of spatial resolution $64 \times 64$.\par
\noindent \textbf{`Shape Residual'}: Same as `Albedo Residual' (weights are not shared).\par
\noindent \textbf{`Albedo Conv.'}: Up - C128(k3) - C64(k3) - C*3(k1)\\
`Up' is bilinear up-sampling that converts spatial resolution from $64 \times 64$ to $128 \times 128$. `CN(kS)' represents 2D convolutional layers with N $S \times S$ filters with stride 1, followed by Batch Normalization and LReLU. `C*N(kS)' represents only 2D convolutional layer with N $S \times S$ filters with stride 1. The output of `Albedo Conv.' produces a tensor of spatial resolution $128 \times 128$ and is RGB channels of albedo.\par
\noindent \textbf{`Shape Conv.'}: Similar to `Albedo Conv.' (weights are not shared), except the last layer `C*3(k1)' is replaced by `C*12(k1)' to produce 12 channels. The 1st to 3rd channels are surface normal, and the 4th to the 12th channels are spherical harmonic coefficients of visibility.\par
\noindent \textbf{`Illumination Estimator'}: It first concatenates the output of `Conv.', `Albedo Residual' and `Shape Residual' to produce a tensor of $128 \times 3 = 384$ channels. This is further processed by 128 $1 \times 1$ 2D convolution, Batch Normalization, LReLU, followed by Average Pooling over $64 \times 64$ spatial resolution to produce 128-dimensional features. The 128-dimensional feature is passed through a fully connected layer to produce 27-dimensional spherical harmonic coefficients of illumination (9 for each RGB channels).\par
We normalize the surface normal of `Shape Conv.' to unit length and append sigmoid activation function to the output of `Albedo Conv.' to ensure albedo is in the range of $[0, 1]$.

\subsection{Training Details}
As specified in Section \ref{sec:trainingoverview}, the training consists of two stages. In the first stage, we train on synthetic data with a mini-batch size of 20. The reconstruction loss $E_{recon}$ is composed of 4 univariate losses $E_{recon-\rho}$, $E_{recon-n}$, $E_{recon-v}$ and $E_{recon-L}$. The name univariate means there is only one variable. For example $E_{recon-\rho}$ is the error between the input image $\hat{I}$ and image rendered by $f_{render}(\rho, \hat{n}, \hat{v}, \hat{L})$, where $\hat{n}$, $\hat{v}$ and $\hat{L}$ are the ground truth of normal, visibility and illumination respectively. We only use univariate reconstruction losses in the first stage because we find by experiments that multivariate reconstruction losses are unstable to train. Our loss function in the first stage is:
\begin{equation}\label{eq:first_stage_loss}
\begin{aligned}
E = & \ E_{recon-\rho} + \lambda_n E_{recon-n} \\
+ & \lambda_v E_{recon-v} + \lambda_L E_{recon-L} \\
+ & \ E_\rho + \lambda_n E_n + \lambda_v E_v + \lambda_L E_L.
\end{aligned}	
\end{equation}
Then we run the trained network on the CelebA dataset \cite{8} to generate pseudo ground truth labels for the second stage.\par
In the second stage, we train with a mini-batch size of 40, in which 20 are synthetic data, and the other 20 are real-world data with pseudo ground truth labels. The loss for the synthetic part is the same as in equation (\ref{eq:first_stage_loss}). Since the pseudo ground truth labels generated in the first stage are biased, we don't use univariate reconstruction losses for the real-world part. Instead, we use three-variate reconstruction loss $E_{recon-\rho,n,L}$. Our loss function for the real world part in the second stage is:
\begin{equation}
E = E_{recon-\rho,n,L} + E_\rho + \lambda_n E_n + \lambda_v E_v + \lambda_L E_L.
\end{equation}
The three-variate reconstruction loss $E_{recon-\rho,n,L}$ is derived from four-variate reconstruction $f_{render}(\rho, n, v, L)$ by treating the visibility as constant and do not backpropagate gradients for it. We block the visibility gradients because we found visibility tends to overfit dark regions for real-world data.

Using different reconstruction losses for synthetic and real-world data stabilizes training and prevents trivial solutions for real-world data. We show in Section \ref{sec:inverse_rendering_results} that our `pseudo supervision' training achieves better reconstruction quality than `Relighting Humans' \cite{6} on the real-world test data they provide.
\par
We use Tensorflow \cite{35} to implement the neural networks. The number of images in the CelebA dataset we used for training is about 170,000. We use Adam solver \cite{36} with the learning rate set to be $10^{-4}$ and mean squared error metric for all losses. We set the balancing loss weights $\lambda_n=\lambda_v=0.2$ and $\lambda_L=0.01$. We uniformly sample 64,000 directions on the sphere to compute the illumination loss $E_L$. In the first stage, we trained 194,209 steps in 5 days. In the second stage, we trained 102,177 steps in another 5 days. All experiments were conducted on a desktop PC with E3-1230 v3@3.30GHz CPU, 32GB DDR3 RAM, and an NVIDIA GTX 1080 8GB GPU.

\section{Comparison with State-of-the-Art Methods}
First, we compare our method with `Relighting Humans' \cite{6} on the data they provide. The target is to show that our method can recover both surface normal and visibility, resulting in better performance on reconstructing real-world images.
Then, we qualitatively compare our inverse rendering result with SfSNet \cite{2} on the data they provide and on the CelebA dataset \cite{8} to show that our method better recovers the surface normal details and produces better shading where visibility effect is strong. Finally, we compare recovered surface normals quantitatively with SfSNet on the Photoface dataset \cite{7} to show our method's correctness. In these comparisons, we use the code released by SfSNet and `Relighting Humans' with no modification.

\subsection{Evaluation of Inverse Rendering}\label{sec:inverse_rendering_results}
\begin{center}
	\captionsetup[subfigure]{labelformat=empty}	
	\centering
	\relightinghumans{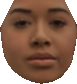}\vspace{0.5mm}
	
	\our{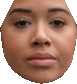}\vspace{0.5mm}
	
	\relightinghumans{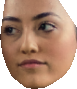}\vspace{0.5mm}
	
	\our{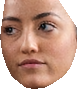}\vspace{0.5mm}
	
	\relightinghumans{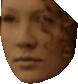}\vspace{0.5mm}
	
	\our{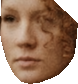}\vspace{0.5mm}
	
	\relightinghumans{1}\vspace{0.5mm}
	
	\our{1}\vspace{0.5mm}
	
	\relightinghumans{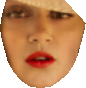}\vspace{0.5mm}
	
	\our{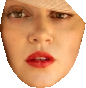}
	
	\componentcaption
	
	\vspace{2mm}
	\parbox[c]{8.3cm}{\footnotesize{Fig.3.~}  Inverse rendering results of `Relighting Humans' (ReH) \cite{6} and our method. The images are provided by ReH. Both ReH and our method are not trained on these images. Our method achieves better reconstruction (Recon.) quality since we are able to train on real-world images using reconstruction loss.}
\end{center}

\vspace{1mm}

We compare inverse rendering results of our method with that of `Relighting Humans' (ReH) \cite{6}. The examples shown in Fig.3 are provided by ReH. Both ReH and our method are not trained on these images. 
Note that the input of ReH is entire human body, we apply their method on the entire human body image and crop the face region as the input to our method. 
Since ReH directly estimates the product of cosine and visibility functions, it does not recover surface normals, while our method is able to recover surface normals. 
In addition, since we can use reconstruction loss to train on real-world images while ReH cannot, the result shows that our method performs better on reconstruction quality. In ReH, they reported they failed to use multivariate reconstruction losses for training on real-world data. They thought it is because their inferred light transfer function has much larger degrees of freedom than surface normal and thus more difficult to fine-tune. While we propose to use different reconstruction losses for synthetic and real-world data, so we can fine-tune on real-world data to achieve better reconstruction quality.

In comparison with SfSNet \cite{2}, we show inverse rendering results on the examples provided by SfSNet in Fig.4, and on the CelebA dataset in Fig.5. We show illumination transfer results compared with SfSNet on the CelebA dataset in Fig.6, where the illumination of the source image is used to replace the illumination of the target image. Note the correct shading should be dark around the nostrils and oral cavity since the visibility effect is strong in these areas. However, the shading of SfSNet in these areas is over smoothed, resulting in the incorrectly baked albedo and visibility effects. At the same time, our method better reveals surface normal and shading details for these areas. We show more inverse rendering results on dark skin persons in Fig.7. The result shows that our visibility estimation is robust to skin tone. More inverse rendering and illumination transfer results on the CelebA dataset are provided in Online Resource 1.

\vfill\null
\columnbreak

\begin{center}
	\captionsetup[subfigure]{labelformat=empty}
	\centering
	\sfsnet{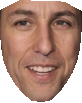}\vspace{0.5mm}
	
	\our{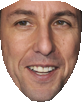}\vspace{0.5mm}
	
	\sfsnet{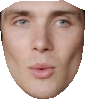}\vspace{0.5mm}
	
	\our{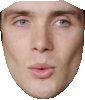}\vspace{0.5mm}
	
	\sfsnet{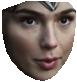}\vspace{0.5mm}
	
	\our{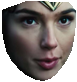}\vspace{0.5mm}
	
	\sfsnet{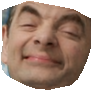}\vspace{0.5mm}
	
	\our{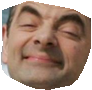}\vspace{0.5mm}
	
	\sfsnet{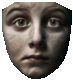}\vspace{0.5mm}
	
	\our{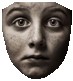}\vspace{0.5mm}
	
	\sfsnet{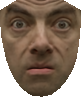}\vspace{0.5mm}
	
	\our{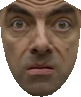}
	
	\componentcaption
	
	\vspace{2mm}
	\parbox[c]{8.3cm}{\footnotesize{Fig.4.~} Inverse rendering results compared with SfSNet on the examples provided by SfSNet \cite{2}. Our method better reveals surface normal and shading details where visibility effect is strong.}
\end{center}

\vfill\null
\columnbreak

\begin{center}
	\captionsetup[subfigure]{labelformat=empty}
	\centering
	\sfsnet{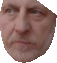}\vspace{0.5mm}
	
	\our{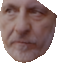}\vspace{0.5mm}
	
	\sfsnet{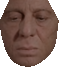}\vspace{0.5mm}
	
	\our{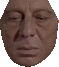}\vspace{0.5mm}
	
	\sfsnet{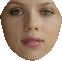}\vspace{0.5mm}
	
	\our{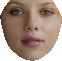}\vspace{0.5mm}
	
	\sfsnet{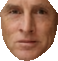}\vspace{0.5mm}
	
	\our{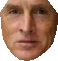}\vspace{0.5mm}
	
	\sfsnet{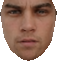}\vspace{0.5mm}
	
	\our{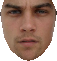}
	
	\componentcaption
	
	\vspace{2mm}
	\parbox[c]{8.3cm}{\footnotesize{Fig.5.~} Inverse rendering results compared with SfSNet \cite{2} on the CelebA dataset. Our method better reveals surface normal and shading details where visibility effect is strong.}
\end{center}

\vfill\null
\columnbreak

\begin{center}
	\captionsetup[subfigure]{labelformat=empty}
	\centering
	\sfsnettransfer{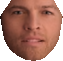}\vspace{0.5mm}
	
	\ourtransfer{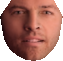}\vspace{0.5mm}
	
	\sfsnettransfer{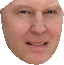}\vspace{0.5mm}
	
	\ourtransfer{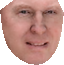}\vspace{0.5mm}
	
	\sfsnettransfer{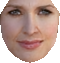}\vspace{0.5mm}
	
	\ourtransfer{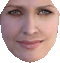}\vspace{0.5mm}
	
	\sfsnettransfer{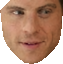}\vspace{0.5mm}
	
	\ourtransfer{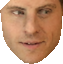}\vspace{0.5mm}
	
	\sfsnettransfer{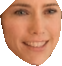}\vspace{0.5mm}
	
	\ourtransfer{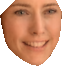}
	
	\vspace{1mm}
	\begin{minipage}{\textwidth}
		\footnotesize{\quad Source \; S-Source \, S-Trans. Transfer \, S-Target \; Target}
	\end{minipage}
	
	\vspace{2mm}
	\parbox[c]{8.3cm}{\footnotesize{Fig.6.~} Illumination transfer results compared with SfSNet \cite{2} on the CelebA dataset. The illumination of the source image is used to replace the illumination of the target image. `S-' denotes shading and `Trans.' denotes transfer.}
	
\end{center}

\vfill\null
\columnbreak

\begin{center}
	\captionsetup[subfigure]{labelformat=empty}
	\centering
	\sfsnet{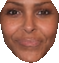}\vspace{0.5mm}
	
	\our{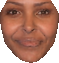}\vspace{0.5mm}
	
	\sfsnet{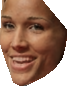}\vspace{0.5mm}
	
	\our{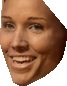}\vspace{0.5mm}
	
	\sfsnet{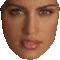}\vspace{0.5mm}
	
	\our{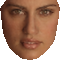}\vspace{0.5mm}
	
	\sfsnet{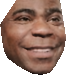}\vspace{0.5mm}
	
	\our{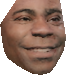}\vspace{0.5mm}
	
	\sfsnet{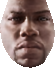}\vspace{0.5mm}
	
	\our{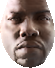}\vspace{0.5mm}
	
	\sfsnet{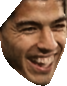}\vspace{0.5mm}
	
	\our{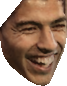}\vspace{0.5mm}
	
	\sfsnet{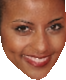}\vspace{0.5mm}
	
	\our{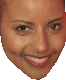}
	
	\componentcaption
	
	\vspace{2mm}
	\parbox[c]{8.3cm}{\footnotesize{Fig.7.~} Inverse rendering results compared with SfSNet \cite{2} on dark skin persons in the CelebA dataset. The results show that our visibility estimation is robust to skin tone.}
\end{center}

\subsection{Evaluation of Surface Normal Recovery}
We compare our reconstructed surface normals with those recovered by the state-of-the-art method SfSNet \cite{2} on the Photoface dataset \cite{7}. The Photoface dataset consists of 3174 sessions of 453 people. We collect all frontal facing images whose face region is detectable, which gives 7665 images. The metric used for this task is the mean angular error of the normals and the percentage of correct pixels below various angular error thresholds. Both SfSNet and our method are not trained on this dataset. The result in Table 1 shows that our method performs as good as SfSNet.

\tabcolsep 8.2pt
\renewcommand\arraystretch{1.3}
\begin{center}
	{\footnotesize{\bf Table 1.} Surface Normal Error on the Photoface Dataset}\\
	\vspace{2mm}
	\footnotesize{
		\begin{tabular*}{\linewidth}{ccccc}\hline\hline\hline
			Algorithm & Mean $\pm$std& $< 20\degree$ & $< 25\degree$ & $< 30\degree$\\\hline
			SfSNet\cite{2} &23.6 $\pm$9.3& 48.3\%&61.3\%&71.4\%\\
			Our&\textbf{23.1 $\pm$9.1}&\textbf{48.4}\%&\textbf{61.5\%}&\textbf{71.8\%}
			\\\hline\hline\hline
		\end{tabular*}
		\\\vspace{1mm}\parbox{8.3cm}{Note: Surface normal reconstruction error on the Photoface dataset\cite{7}. SfSNet \cite{2} and our method are not trained on this dataset. Lower is better for mean angular error in degree (column 2), and higher is better for the percentage of correct pixels below various thresholds (columns 3 to 5).}
	}
\end{center}

\section{Conclusions}
In this paper, we investigate the simultaneous estimation of the surface normal and visibility of a human face image. Compared with the previous method `Relighting Humans' \cite{6}, we separately estimate the visibility and cosine term related to surface normal, which allows us to recover the surface normal. We propose a more stable reconstruction loss for training on real-world images, even with large degrees of freedom in the visibility term. Thus our method achieves better reconstruction quality than \cite{6} on the real-world test data they provide. We quantitatively and qualitatively show the correctness of our method on surface normal recovery. We also show that our method better recovers the surface normal details and produces better shading where the visibility effect is strong.

Our method also has limitations. Since our synthetic training data doesn't contain beard, our estimated visibility tends to overfit in hairy regions, as shown in Fig.8. For future work, we believe more accurate modeling of the rendering process, such as glossy reflection caused by the oil layer on human faces, subsurface scattering of human face skins, and simulation of global illumination \cite{37} need to be addressed.

\begin{center}
	\captionsetup[subfigure]{labelformat=empty}
	\centering
	\beard{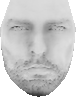}
	
	\beard{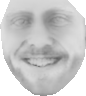}
	
	\beard{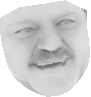}
	
	\vspace{1mm}
	\footnotesize{\;Input \quad Albedo \; Visibility}
	
	\vspace{2mm}
	\parbox[c]{8.3cm}{\footnotesize{Fig.8.~} Limitations of our method, our visibility function overfitted to dark albedo regions such as beard.}
\end{center}

\appendix
\renewcommand{\thesection}{A.\arabic{section}}
\setcounter{section}{0}

\section{Spherical Harmonic}
The Laplace spherical harmonics $Y_l^m(\omega)$ form a complete set of orthonormal functions and thus form an orthonormal basis of the Hilbert space of square-integrable functions. On the unit sphere, any square-integrable function $f(\omega)$ can thus be projected onto these bases and be expanded as a linear combination of these:
\begin{equation}
f(\omega) = \sum_{l=0}^{\infty}\sum_{m=-l}^{l}f_l^m Y_l^m(\omega)
\end{equation}
\begin{equation}\label{eq:sh_projection}
f_l^m = \int_{\Omega} f(\omega)Y_l^m(\omega) d\omega
\end{equation}
where $\Omega$ denotes the sphere, $l \in \N$ and $m \in \Z$ are the degree and order of the basis respectively.
We use right-hand coordinate system convention, a unit length vector $\omega=(x,y,z)^T$ in Cartesian coordinate can be expressed in spherical coordinate:
\begin{equation}
\begin{aligned}
x &= \sin(\theta)\cos(\phi)\\
y &= \sin(\theta)\sin(\phi)\\
z &= \cos(\theta)
\end{aligned}
\end{equation}
where $\theta \in [0, \pi]$ denotes colatitude angle and $\phi \in [0,2\pi)$ denotes azimuth angle.
\par
It has been shown in \cite{38} that cosine function defined on the upper hemisphere in equation (\ref{eq:f_render}) can be well approximated by $l \leq 2$:
\begin{equation}
(Y_0^0, Y_1^{-1}, Y_1^0, Y_1^1, Y_2^{-2}, Y_2^{-1}, Y_2^0, Y_2^1, Y_2^2)^T
\end{equation}
where
\begin{equation}\label{eq:sh_basis}
\begin{alignedat}{2}
&Y_0^0    = \frac{1}{2}\sqrt{\frac{1}{\pi}}
&&Y_1^{-1} = \sqrt{\frac{3}{4\pi}}y\\
&Y_1^0    = \sqrt{\frac{3}{4\pi}}z
&&Y_1^1    = \sqrt{\frac{3}{4\pi}}x\\
&Y_2^{-2} = \frac{1}{2}\sqrt{\frac{15}{\pi}}xy
&&Y_2^{-1} = \frac{1}{2}\sqrt{\frac{15}{\pi}}yz\\
&Y_2^0    = \frac{1}{4}\sqrt{\frac{5}{\pi}}(3z^2-1)\qquad
&&Y_2^1    = \frac{1}{2}\sqrt{\frac{15}{\pi}}xz\\
&Y_2^2    = \frac{1}{4}\sqrt{\frac{15}{\pi}}(x^2-y^2)
\end{alignedat}
\end{equation}
For clarity, let $i=l(l+1)+m+1$ and we have $Y_l^m = Y_i$. 

\subsection{Rotation}
Spherical harmonics can be rotated, let $g(\omega)$ denotes the rotated version of $f(\omega)$:
\begin{equation}
g(R\omega) = f(\omega)
\end{equation}
where $R \in SO(3)$ is a rotation. Let $g_i$ and $f_i$ denote their spherical harmonic coefficients respectively:
\begin{equation}
f(\omega) = \sum_{i} f_i Y_i(\omega)
\end{equation}
\begin{equation}
g(\omega) = \sum_{i} g_i Y_i(\omega)
\end{equation}
in which $g_i$ can be expressed as a linear combination of $f_i$:
\begin{equation}\label{eq:sh_rotation}
g_i = \sum_{j} f_j M_{ij}
\end{equation}
where $M_{ij}$ is:
\begin{equation}\label{eq:sh_rotation_matrix_definition}
M_{ij} = \int_{\Omega}{Y_i(R\omega)Y_j(\omega) d\omega}
\end{equation}
To show equation (\ref{eq:sh_rotation_matrix_definition}) is the case, we start from the definition of $g_i$:
\begin{equation}
\begin{aligned}
g_i &= \int_{\Omega}{Y_i(\omega) g(\omega) d\omega}\\
&= \int_{\Omega}{Y_i(R\omega) g(R\omega) dR\omega}\\
&= \int_{\Omega}{Y_i(R\omega) f(\omega) dR\omega}\\
&= \int_{\Omega}{Y_i(R\omega) \Big[\sum_{j}{f_j Y_j(\omega)}\Big] dR\omega}\\
&= \sum_{j} f_j \int_{\Omega}{Y_i(R\omega) Y_j(\omega) dR\omega}\\
&= \sum_{j} f_j \int_{\Omega}{Y_i(R\omega) Y_j(\omega) d\omega}\\
&= \sum_{j} f_j M_{ij}\\
\end{aligned}
\end{equation}
With Proper Euler angle ZYZ formulation, a rotation $R(\alpha, \beta, \gamma)$ is decomposed into three consecutive extrinsic rotations $Z_\alpha$, $Y_\beta$ and $Z_\gamma$:
\begin{equation}
R(\alpha, \beta, \gamma) = Z_\gamma Y_\beta Z_\alpha
\end{equation}
where $Z_\alpha$ denotes a rotation of $\alpha$ about the Z-axis. $Y_\beta$ can be further decomposed into a rotation of $+90\degree$ about the X-axis, followed by a rotation of $\beta$ about the Z-axis and finally a rotation of $-90\degree$ about the X-axis:
\begin{equation}
R(\alpha, \beta, \gamma) = Z_\gamma X_{-90\degree} Z_\beta X_{+90\degree} Z_\alpha
\end{equation}
With equations (\ref{eq:sh_basis}) and (\ref{eq:sh_rotation_matrix_definition}) we can derive the formulae for $Z_\alpha$, $X_{+90\degree}$ and $X_{-90\degree}$. Since each of them is a $9 \times 9$ sparse matrix, we show only the nonzero elements, where $Z_{ij}$ represents the row $i$ and column $j$ of $Z_\alpha$:
\begin{equation}
\begin{alignedat}{3}
&Z_{11} = 1  &&Z_{22} = \cos{\alpha}  &&Z_{24} = \sin{\alpha}\\
&Z_{33} = 1  &&Z_{42} = -\sin{\alpha}  &&Z_{44} = \cos{\alpha}\\
&Z_{55} = \cos{2\alpha} \quad\; &&Z_{59} = \sin{2\alpha}  &&Z_{66} = \cos{\alpha}\\
&Z_{68} = \sin{\alpha}  &&Z_{77} = 1  &&Z_{86} = -\sin{\alpha}\\
&Z_{88} = \cos{\alpha}  &&Z_{95} = -\sin{2\alpha} \quad\; &&Z_{99} = \cos{2\alpha}
\end{alignedat}
\end{equation}
The equation for $X_{+90\degree}$ is:
\begin{equation}
\begin{alignedat}{3}
&X_{11} = 1 &&X_{23} = -1 &&X_{32} = 1\\
&X_{44} = 1 &&X_{58} = -1 &&X_{66} = -1\\
&X_{77} = -\frac{1}{2}	\qquad\quad	&&X_{79} = -\frac{\sqrt{3}}{2} \qquad\quad &&X_{85} = 1\\
&X_{97} = -\frac{\sqrt{3}}{2} &&X_{99} = \frac{1}{2}
\end{alignedat}
\end{equation}
The equation for $X_{-90\degree}$ is:
\begin{equation}
\begin{alignedat}{3}
&X_{11} = 1 &&X_{23} = 1 &&X_{32} = -1\\
&X_{44} = 1 &&X_{58} = 1 &&X_{66} = -1\\
&X_{77} = -\frac{1}{2}	\qquad\quad	&&X_{79} = -\frac{\sqrt{3}}{2} \qquad\quad &&X_{85} = -1\\
&X_{97} = -\frac{\sqrt{3}}{2} &&X_{99} = \frac{1}{2}
\end{alignedat}
\end{equation}

\subsection{Integration of the Product of Two Functions}	
Since spherical harmonic basis functions are orthonormal:
\begin{equation}
\int_{\Omega}{Y_i(\omega) Y_j(\omega) d\omega} = 
\begin{cases}
1,& \text{if } i = j\\
0,& \text{otherwise}
\end{cases}
\end{equation}
The integration of the product of two functions after projection onto spherical harmonic basis is the dot product of their coefficients:
\begin{equation}\label{eq:two_function_product}
\begin{aligned}
& \quad \: \int_{\Omega}{f(\omega)g(\omega) d\omega}\\
& = \int_{\Omega}{\Big[\sum_{i}{f_i Y_i(\omega)}\Big] \Big[\sum_{j}{g_j Y_j(\omega)}\Big] d\omega}\\
&= \sum_{i}\sum_{j} f_i g_j \int_{\Omega}{ Y_i(\omega)  Y_j(\omega) d\omega}\\
&= \sum_{i}f_i g_i
\end{aligned}
\end{equation}

\subsection{Integration of the Product of Three Functions}
Let $e(\omega)$ denotes the product of $f(\omega)$ and $g(\omega)$:
\begin{equation}
e(\omega) = f(\omega) g(\omega)
\end{equation}
The coefficient $e_i$ of $e(\omega)$ after projection onto spherical harmonic basis is:
\begin{equation}\label{eq:merge_two_sh}
\begin{aligned}
e_i &= \int_{\Omega} Y_i(\omega) e(\omega) d\omega\\
&= \int_{\Omega} Y_i(\omega) [ f(\omega) g(\omega) ] d\omega\\
&= \int_{\Omega} Y_i(\omega) \Big[ \Big( \sum_j f_j Y_j(\omega) \Big) \Big( \sum_k g_k Y_k(\omega) \Big) \Big] d\omega\\
&= \sum_{j} \sum_{k} f_j g_k \int_{\Omega} Y_i(\omega) Y_j(\omega) Y_k(\omega) d\omega\\
&= \sum_{j} \sum_{k} f_j g_k T_{ijk}
\end{aligned}
\end{equation}
where $T_{ijk}$ is called tripling coefficients \cite{39}. The integration of the product of three functions $f(\omega)$, $g(\omega)$ and $h(\omega)$ is:
\begin{equation}\label{eq:triple_product_integration}
\begin{aligned}	
& \quad \: \int_{\Omega}{ h(\omega)f(\omega)g(\omega) d\omega}\\
&= \int_{\Omega}{ h(\omega)e(\omega) d\omega}\\
&= \sum_{i} h_i e_i\\
&= \sum_{i} \Big(h_i \sum_{j} \sum_{k} f_j g_k T_{ijk}\Big)
\end{aligned}
\end{equation}
With equation (\ref{eq:sh_basis}) and (\ref{eq:merge_two_sh}) we can derive the formula for $T_{ijk}$. Note $\{T_{ijk}\}$ are equal under permutation of $i, j, k$ by definition:
\begin{equation}
T_{ijk} = T_{ikj} = T_{jik} = T_{jki} = T_{kij} = T_{kji}
\end{equation}
Since most elements in $T$ are zero, we only show the nonzero $T_{ijk}$ when $i \leq j \leq k$:
\begin{equation}
T_{111} = \sqrt{\frac{1}{4\pi}}
\end{equation}
\begin{equation}
T_{122} = \sqrt{\frac{1}{4\pi}}
\end{equation}
\begin{equation}
T_{133} = \sqrt{\frac{1}{4\pi}}
\end{equation}
\begin{equation}
T_{144} = \sqrt{\frac{1}{4\pi}}
\end{equation}
\begin{equation}
T_{155} = \sqrt{\frac{1}{4\pi}} \qquad T_{245} = \sqrt{\frac{3}{20\pi}}
\end{equation}
\begin{equation}
T_{166} = \sqrt{\frac{1}{4\pi}} \qquad T_{236} = \sqrt{\frac{3}{20\pi}}
\end{equation}
\begin{equation}
\begin{alignedat}{2}
&T_{177} = \sqrt{\frac{1}{4\pi}} &&T_{227} = -\sqrt{\frac{1}{20\pi}}\\
&T_{337} = \sqrt{\frac{1}{5\pi}} &&T_{447} = -\sqrt{\frac{1}{20\pi}}\\
&T_{557} = -\frac{1}{7}\sqrt{\frac{5}{\pi}} \qquad &&T_{667} = \frac{1}{14}\sqrt{\frac{5}{\pi}}\\
&T_{777} = \frac{1}{7}\sqrt{\frac{5}{\pi}}
\end{alignedat}
\end{equation}
\begin{equation}
\begin{alignedat}{2}
&T_{188} = \sqrt{\frac{1}{4\pi}} &&T_{348} = \sqrt{\frac{3}{20\pi}}\\
&T_{668} = \frac{1}{14}\sqrt{\frac{15}{\pi}} \qquad &&T_{788} = \frac{1}{14}\sqrt{\frac{5}{\pi}}
\end{alignedat}
\end{equation}
\begin{equation}
\begin{alignedat}{2}
&T_{199} = \sqrt{\frac{1}{4\pi}} &&T_{229} = -\sqrt{\frac{3}{20\pi}}\\
&T_{449} = \sqrt{\frac{3}{20\pi}} &&T_{669} = -\frac{1}{14}\sqrt{\frac{15}{\pi}}\\
&T_{799} = -\frac{1}{7}\sqrt{\frac{5}{\pi}} \qquad &&T_{889} = \frac{1}{14}\sqrt{\frac{15}{\pi}}
\end{alignedat}
\end{equation}

\section{Recover Illumination from Shading}\label{section:recover_illumination_from_shading}
In order to recover incident illumination coefficients $\{L_i\}_{i=1}^9$ from the shading coefficients $\{S_i\}_{i=1}^9$ in \cite{2, 29}:
\begin{equation}\label{eq:shading}
\begin{aligned}
S(n) &= \int_{\Omega^+} L(\omega) \cos(\omega, n) d\omega\\
&= \sum_{i=1}^{9} S_i Y_i(n)
\end{aligned}
\end{equation}
With equation (\ref{eq:shading}) and (\ref{eq:two_function_product}), we have:
\begin{equation}\label{eq:shading_illumination_cosine}
\sum_{i=1}^{9} S_i Y_i(n) = \sum_{i=1}^{9} L_i c_i
\end{equation}
where $\{L_i\}_{i=1}^9$ are the illumination coefficients we want to recover, $\{c_i\}_{i=1}^9$ are the coefficients of cosine function defined on upper hemisphere in equation (\ref{eq:cosine_expansion}). We establish an over-determined system of linear equations (\ref{eq:shading_illumination_cosine}) by generating 64,000 uniformly distributed directions $\{\omega_i\}_{i=1}^{64,000}$ on sphere to solve for $\{L_i\}$.

\section{Extended Articles}

There are some extended articles about applying physical lighting computation in various applications:

\begin{enumerate}
    \item Deep Learning-Based Monte Carlo Noise Reduction By training a neural network denoiser through offline learning, it can filter noisy Monte Carlo rendering results into high-quality smooth output, greatly improving physics-based Availability of rendering techniques \cite{huo2021survey}, common research includes predicting a filtering kernel based on g-buffer \cite{bako2017kernel}, using GAN to generate more realistic filtering results \cite{xu2019adversarial}, and analyzing path space features Perform manifold contrastive learning to enhance the rendering effect of reflections \cite{cho2021weakly}, use weight sharing to quickly predict the rendering kernel to speed up reconstruction \cite{fan2021real}, filter and reconstruct high-dimensional incident radiation fields for unbiased reconstruction Drawing guide \cite{huo2020adaptive}, etc.
    \item The multi-light rendering framework is an important rendering framework outside the path tracing algorithm. Its basic idea is to simplify the simulation of the complete light path illumination transmission after multiple refraction and reflection to calculate the direct illumination from many virtual light sources, and provide a unified Mathematical framework to speed up this operation \cite{dachsbacher2014scalable}, including how to efficiently process virtual point lights and geometric data in external memory \cite{wang2013gpu}, how to efficiently integrate virtual point lights using sparse matrices and compressed sensing \cite{huo2015matrix}, and how to handle virtual line light data in translucent media \cite{huo2016adaptive}, use spherical Gaussian virtual point lights to approximate indirect reflections on glossy surfaces \cite{huo2020spherical}, and more.
    \item Automatic optimization of drawing pipelines Apply high-quality drawing technology to real-time drawing applications by optimizing drawing pipelines. The research contents include automatic optimization based on quality and speed \cite{wang2014automatic}, automatic optimization for energy saving \cite{ wang2016real,zhang2021powernet}, LOD optimization for terrain data \cite{li2021multi}, automatic optimization and fitting of pipeline drawing signals \cite{li2020automatic}, anti-aliasing \cite{zhong2022morphological}, etc.
    \item Using physically-based process to guide the generation of data for single image reflection removal \cite{kim2020single}; propagating local image features in a hypergraph for image retreival \cite{an2021hypergraph}; managing 3D assets in a block chain-based distributed system \cite{park2021meshchain}.
\end{enumerate}

\bibliographystyle{ieee}
\bibliography{srbib}

\label{last-page}
\end{multicols}
\label{last-page}
\end{document}